\documentclass{article}
\usepackage[preprint,nonatbib]{neurips_2026}
\usepackage[utf8]{inputenc}
\usepackage[T1]{fontenc}
\usepackage{courier}
\usepackage{hyperref}
\usepackage{caption}
\usepackage{graphicx}
\usepackage[export]{adjustbox}
\graphicspath{ {./images/} }
\usepackage{amsmath}
\usepackage{amsfonts}
\usepackage{amssymb}
\usepackage[version=4]{mhchem}
\usepackage{stmaryrd}
\usepackage{booktabs}
\usepackage{array}
\usepackage{url}
\usepackage{nicefrac}
\usepackage{microtype}
\usepackage{xcolor}
\usepackage{algorithm}
\usepackage{algpseudocode}
\title{Formal Skill: Programmable Runtime Skills for Efficient and Accurate LLM Agents}

\author{%
Xi Zhang
\And
Meijun Gao
\And
Yuntian Zhao
\And
Xinyu Tan
\And
Yilun Yao
\AND
Feiyu Wang
\And
Yanshu Wang
\And
Dingsiyi
\And
Tong Yang%
}
\date{}

\DeclareUnicodeCharacter{2192}{\ifmmode\rightarrow\else{$\rightarrow$}\fi}
\DeclareUnicodeCharacter{2713}{\ifmmode\checkmark\else{$\checkmark$}\fi}
\DeclareUnicodeCharacter{00D7}{\ifmmode\times\else{$\times$}\fi}

\begin{document}
\maketitle
\captionsetup{singlelinecheck=false}

\begin{abstract}
Large Language Model (LLM) agents increasingly act inside real workspaces, where tools and skills determine whether model reasoning becomes reliable action. Existing skills remain largely informal: Markdown skills and instruction packs encode procedures as long natural-language documents, while function calling, Model Context Protocol (MCP) servers, and framework tools structure individual actions but usually leave workflow state, policy enforcement, and completion discipline outside the skill itself. We introduce \textbf{Formal Skill}, a runtime-native abstraction that represents reusable capability with JSON metadata and action schemas, reliable Python executors, hook-governed control logic, Formal Skill routing, and skill-local runtime state. By moving reusable procedure from repeated prompt text into executable state machines and hook policies, Formal Skill gives agents a token-efficient and enforceable control surface. We implement the abstraction in FairyClaw, an open-source event-driven runtime for executable, observable, and composable Formal Skills. On Harness-Bench, FairyClaw obtains highly competitive average scores while using substantially fewer tokens, with especially strong results on tasks that expose the role of Formal Skill.
\end{abstract}

\section{Introduction}

Large Language Model (LLM) agents are rapidly becoming a central interface for software engineering, office automation, research assistance, and operational decision making. Systems such as Claude Code \cite{claude_code_skills}, OpenClaw \cite{openclaw_runtime}, and Hermes \cite{hermes_docs} illustrate a shift from chat-oriented assistants toward agents that inspect repositories, run commands, edit files, call external services, and coordinate multi-step work over time. In these agents, skills and tools are not peripheral conveniences: they are the agent's hands and feet. They determine whether a language model can move from describing the world to acting on it.

Existing systems organize reusable agent capability mainly in two ways. First, \emph{text-packaged procedural skills} such as Claude-style skills and many \texttt{SKILL.md} libraries bundle discovery metadata, Markdown procedures, auxiliary files, and optional scripts; tools may be invoked inside the procedure, but the procedure itself is primarily a document read by the model. Second, \emph{schema-level tool interfaces} such as function calling, Model Context Protocol (MCP) tools, framework tools, and API-oriented tool-use methods describe individual actions with names, schemas, and protocol boundaries; they make tool invocation more structured, but usually do not define a skill-level workflow that owns phase order, recovery, and completion. Agent runtimes, workspaces, sandboxes, and benchmark harnesses provide the surrounding execution and evaluation environment for both styles, but they do not by themselves make the reusable procedure formal. By \emph{Informal Skills}, we mean skills whose operational semantics live mainly in natural-language instructions, naming conventions, prompt discipline, or external orchestration rather than in a runtime-executable object; across existing approaches, the core limitation is that the skill remains informal in this sense. Informal Skills are costly because long procedural text must be loaded as tokens; ambiguous because natural language remains only semi-structured to the machine; weak because ordering constraints, safety rules, and completion criteria are soft suggestions; and brittle because recovery state is usually implicit in the transcript rather than represented as skill-local runtime state.

To address these limitations, we propose \textbf{Formal Skill} as a runtime-native alternative. A Formal Skill converts a reusable agent capability from a purely natural-language artifact into a structured executable object with five components: (1) JSON metadata and action schemas that define the model-visible interface, (2) reliable executors that implement actions with deterministic validation and bounded side effects, (3) lifecycle hooks that mediate model calls and tool calls, (4) skill-local runtime state that records phase, evidence, gates, and recovery context, and (5) routing metadata that determines when the skill should be exposed to a task or subtask. It is formal because JSON metadata and schemas describe the skill's machine-readable surface, while transition points, state variables, side-effect boundaries, and completion gates are represented as runtime data and executable programs rather than as prose alone. The goal is not to remove language from agent behavior, but to assign it the right role: the model reasons and chooses among structured actions, while the runtime owns the procedural invariants. Token-heavy manuals become compact state-conditioned guidance; ambiguous instructions become schemas and hook policies; soft constraints become executable checks; and implicit recovery becomes an observable state machine.

We build \textbf{FairyClaw} to realize and evaluate the Formal Skill concept. FairyClaw is an event-driven LLM-agent runtime with a single-step planner, typed history Intermediate Representation (IR), capability registry, hook pipeline, persistent session state, sub-agent delegation, and Formal Skill routing. It is therefore not merely a system that uses Formal Skill, but the concrete runtime substrate developed to make Formal Skills routable, executable, observable, and enforceable. We evaluate FairyClaw on the independent Harness-Bench suite \cite{harnessbench}. Among six evaluated agent harnesses, FairyClaw obtains an average overall score of 0.690, ranking third by mean score across two model families and first in the \texttt{gpt-5.4} group, while using 7.35M tokens in total, or 49.0K tokens per task on average. This per-task token use is about 48\% lower than the mean of the other five harnesses and about 33\% lower than the next most token-efficient harness. On tasks that demonstrate typical Formal Skill effects, FairyClaw achieves especially strong task-level results.

Our contributions are:

\begin{itemize}
  \item We introduce \textbf{Formal Skill}, a token-efficient and enforceable skill abstraction that moves procedural knowledge from long natural-language prompts into JSON metadata and action schemas, reliable Python executors, hook-governed control loops, Formal Skill routing, and skill-local runtime state.
  \item We implement \textbf{FairyClaw}, an open-source event-driven agent runtime for Formal Skills, integrating single-step planning, typed history IR, persistent session state, sub-agent delegation, Formal Skill routing, and hook-based execution control. The implementation is available at \url{https://github.com/PKULab1806/FairyClaw}.
\end{itemize}

\section{Related Work}

\subsection{Prompt-based agent skills and \texttt{SKILL.md}}

Prompt-based agent skills are already widely used in practical agent systems. Anthropic's Agent Skills package discovery metadata written in YAML, a required \texttt{SKILL.md} instruction file, and optional scripts or assets under progressive disclosure \cite{anthropic_agent_skills,anthropic_skills_spec,anthropic_skills_blog}; Claude Code adopts this pattern with invocation controls, subagents, allowed tools, plugins, hooks, and MCP integration \cite{claude_code_skills,claude_code_plugins}. These systems correctly identify reusable procedural knowledge as a missing layer between tools and monolithic prompts. Their limitation is that the procedure itself remains mostly text-level: once selected, the Markdown body is still paid for as prompt tokens, interpreted as semi-structured natural language, and trusted to guide ordering, recovery, and completion. Optional scripts can make subroutines deterministic, but the skill-level workflow usually remains prose. Formal Skill preserves the packaging and progressive-disclosure intuition while moving the reusable procedure into JSON metadata and action schemas, reliable executors, lifecycle hooks, and skill-local runtime state.

\subsection{Tool use and structured action interfaces}

Tool-use research shows that structured action interfaces extend model capability: Modular Reasoning, Knowledge and Language (MRKL) routes language models to external modules \cite{mrkl}, ReAct interleaves reasoning with actions and observations \cite{react}, Toolformer learns API-use decisions \cite{toolformer}, and Gorilla, API-Bank, and ToolLLM/ToolBench study API selection and tool-call generation at scale \cite{gorilla,apibank,toolllm}. Production systems instantiate this idea through JSON schemas and protocol boundaries, including OpenAI function calling and Agents Software Development Kit (SDK) tools \cite{openai_function_calling,openai_agents_tools}, MCP tools and resources over JSON Remote Procedure Call (JSON-RPC) \cite{mcp_overview,mcp_tools}, and framework tools such as LangChain \cite{langchain}. These interfaces type individual actions, but a typed action is not yet a formal skill: phase order, recovery state, side-effect policy, and completion gates are typically left to prompts or outer orchestration. Formal Skill is complementary because it organizes structured actions under routing, hooks, skill-local runtime state, and gates so that a tool set becomes an enforceable procedure.

\subsection{Agent runtimes and software engineering agents}

Agent and coding runtimes demonstrate that performance depends on the action space and feedback channel, not only on the base model. Reflexion, Self-Refine, and Voyager study feedback and reusable executable skills \cite{reflexion,selfrefine,voyager}; SWE-agent and AutoCodeRover show that repository navigation, editing, testing, search, and repair workflows are shaped by the agent-computer interface \cite{sweagent,autocoderover}. Engineering systems such as Claude Code \cite{claude_code_skills,claude_code_custom_tools}, Codex CLI \cite{codex_cli,codex_sandbox}, OpenClaw \cite{openclaw_runtime,openclaw_tools}, and Hermes Agent \cite{hermes_docs,hermes_architecture,hermes_envs} expose workspaces, shell and file operations, sandbox or approval policies, MCP integration, subagents, memory, toolsets, and skills. They make agents practically useful, but reusable capability is still usually represented as either individual tools or textual skills. Formal Skill makes the reusable procedure itself a first-class runtime object with compact descriptions, executable policy, lifecycle control, and observable state.

\section{Formal Skill Design and Implementation}

Formal Skill is a runtime contract for reusable agent capability. It is motivated by the same failure modes identified in the introduction and related work: prompt skills encode procedures as expensive semi-structured text, leave execution constraints as soft instructions, and lack explicit skill-local runtime state for recovery and completion. This section defines the abstraction, describes its implementation in FairyClaw, and then instantiates it with \textsc{CodeRepairOps}, a concrete code-repair skill.

\subsection{Failure modes of prompt skills}

A prompt skill is easy to author because it resembles an onboarding document for a human operator. That convenience becomes a liability when the reader is a model whose actions affect a workspace. The problem is not that natural language instructions are useless; it is that they are a weak boundary between reasoning and side effects.

First, prompt skills are \textbf{textual procedures}. A \texttt{SKILL.md} file can describe ordering constraints, forbidden actions, recovery strategies, and reporting requirements, but all of this material must be loaded as tokens and interpreted from prose. Even with progressive disclosure, the loaded instruction is still semi-structured from the runtime's perspective: the system can store the text, but it cannot directly execute the procedure encoded in it.

Second, prompt skills provide \textbf{weak execution constraints}. A paragraph can say that the agent must reproduce a failure before patching, must not edit tests, or must verify before finishing. Unless these constraints are reimplemented in code, however, the runtime cannot hide inappropriate tools, reject unsafe arguments, or block a premature final answer. The model may follow the instruction, but the skill itself does not enforce it.

Third, prompt skills have \textbf{implicit state and recovery}. After a failed command, an invalid patch, or a partially written artifact, the transcript contains evidence of what happened, but the skill has no compact execution object that says which phase it is in, what obligations remain, or why completion is blocked. This encourages ad hoc recovery and self-defined checkpoints, especially in long-horizon tasks.

\subsection{Formal Skill as an executable contract}

A Formal Skill replaces the text-only procedure with a skill-local protocol that the runtime can inspect and enforce. The model still reasons about the task and chooses actions, but the action surface is governed by code, schema, state, and hooks. Its structure follows directly from the three failures above.

\begin{itemize}
  \item \textbf{Structured skill interfaces}: to address text-heavy procedures, a Formal Skill moves most procedural knowledge out of prompt prose and into JSON metadata and action schemas, Python executors, hook code, and configuration. The model sees short tool descriptions and phase guidance, while detailed policies remain executable runtime objects. This reduces repeated token consumption and avoids paying for long natural-language manuals on every relevant turn.
  \item \textbf{Executable enforcement}: to address weak execution constraints, a Formal Skill turns suggestions into runtime checks. Executors validate arguments and side effects; hooks filter visible tools, reject unsafe calls, and prevent completion when verification or artifact gates remain open. Constraints therefore become part of the agent's action semantics rather than instructions the model may or may not follow.
  \item \textbf{Skill-local runtime state}: to address implicit state and recovery, a Formal Skill stores progress in skill-local runtime state consumed by hooks and tools. Fields such as phase, verification status, produced artifacts, required artifacts, and gate-failure reasons make recovery explicit: the runtime can decide whether to continue collecting evidence, return to patching, enter reporting, or block finalization.
\end{itemize}

The resulting abstraction is not merely a more disciplined prompt. It is a division of labor: natural language remains useful for task-specific reasoning, but reusable procedure is represented as compact interfaces, executable policy, and stateful control.

This division also exposes a real trade-off. Prompt skills and \texttt{Skill.md} files are attractive partly because they are easy for non-programmers to author: a user can describe a procedure in natural language without writing schemas, executors, state machines, or hook code. Formal Skill raises the authoring bar because the skill creator must specify JSON metadata, implement reliable Python actions, and sometimes design state transitions enforced by hooks. We view this as a practical cost rather than a conceptual objection. Modern LLMs are already strong code-generation systems, so an informal skill can itself become the source artifact from which an agent synthesizes the corresponding formal components: action schemas, executor skeletons, hook programs, and state definitions. In this sense, Formal Skill can support a bootstrapping path in which users still begin with a human-readable \texttt{Skill.md}, while the runtime or an assisting model compiles it into a token-efficient and enforceable skill representation.

\subsection{Hook-governed programming model}

In FairyClaw, a Formal Skill is packaged as a capability plugin. The manifest declares tools and hooks; configuration files define deterministic policy parameters; scripts implement executors and hook logic. The programming model is intentionally divided along runtime boundaries.

The action plane is where model-visible verbs are defined. A code-repair skill, for example, should not expose an unstructured ``please fix the code'' operation. It should expose smaller actions such as collecting evidence, applying a unified patch, running verification, and writing artifacts. The JSON side tells the model what operation exists and what arguments are valid; the Python side decides what actually happens, including path checks, command allowlists, patch-shape validation, and structured output.

\begin{figure}[htbp]
\centering
\includegraphics[width=0.92\linewidth]{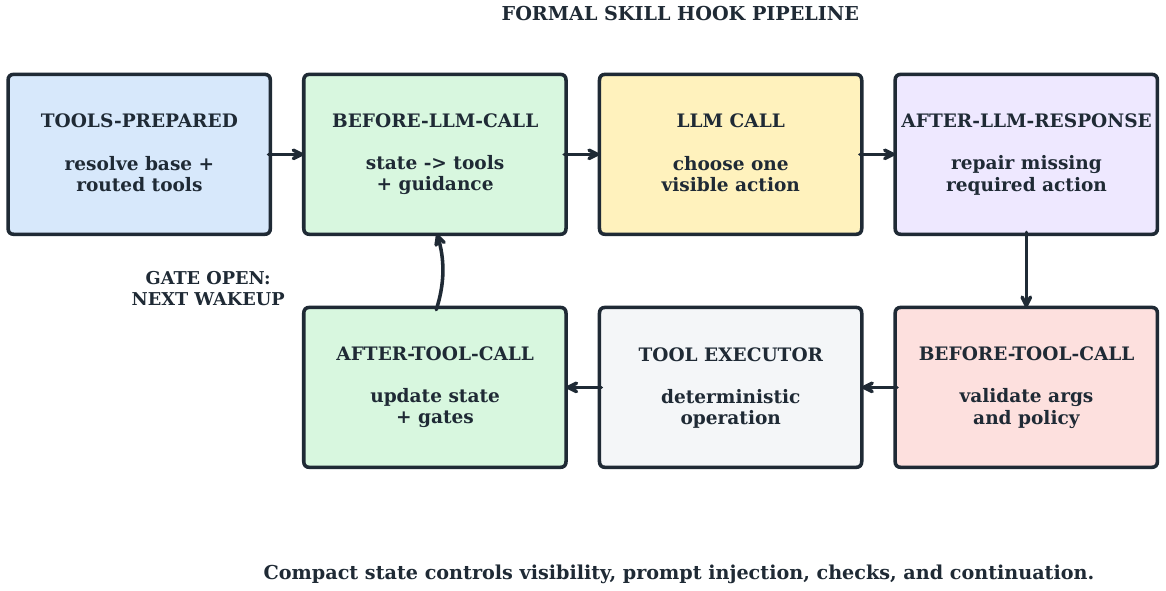}
\caption{Formal Skill hook pipeline. Hooks convert compact skill-local runtime state into tool visibility, prompt guidance, policy checks, and continuation decisions.}
\label{fig:hook-pipeline}
\end{figure}

The control plane is implemented by hooks around the LLM and tool boundary. Figure~\ref{fig:hook-pipeline} shows the pipeline. The point is not that each hook is complex; the point is that the runtime has named intervention points where a skill can turn state into visible tools, guidance, validation, and continuation.

The state plane makes the procedure an execution object rather than a transcript convention. A repair state may include \texttt{phase}, \texttt{verification\_passed}, \texttt{required\_artifacts}, \texttt{produced\_artifacts}, \texttt{gate\_fail\_reasons}, and last-tool status. Hooks consume this state to determine which tools are visible, whether another LLM turn is required, and whether the task can finish.

\subsection{Implementation of Formal Skill in FairyClaw}

FairyClaw is the runtime we build to make Formal Skill executable. Its role is not to add another long system prompt, but to provide the concrete runtime services that a Formal Skill requires: plugin loading, capability routing, hook execution, persistent session state, typed history, workspace scoping, and parent-child task collaboration. A Formal Skill is therefore not embedded as an instruction file inside a prompt. It is registered as a capability plugin whose tools, hooks, configuration, and state are resolved by the runtime before each model decision.

The most important implementation mechanism is skill routing. FairyClaw separates always-available orchestration tools from domain-specific Formal Skills. The parent planner can create a sub-session with a narrower task, workspace, deliverable contract, and routed skill set; the sub-session then sees only the tools and hook policies required for that work. This reduces token cost and action ambiguity at the same time: the model receives fewer irrelevant tool descriptions, and the enabled skill can inject only phase-local guidance rather than a full procedural manual.

\begin{algorithm}[H]
\caption{Sub-agent Formal Skill routing}
\label{alg:skill-routing}
\footnotesize
\begin{algorithmic}[1]
\Require Parent session $p$ and delegated task $d$
\State Read $d$'s task type, workspace root, \texttt{done\_when}, and task text
\State Select candidate Formal Skills from the capability registry
\ForAll{candidate skill $c$}
  \State Score $c$ by task type, trigger metadata, and policy constraints
\EndFor
\State Create sub-session $s$ with workspace root and \texttt{done\_when} copied from $d$
\State Attach selected manifests, configs, hooks, and tool schemas to $s$
\State Expose default orchestration tools plus routed skill tools
\State Run routed hooks around LLM and tool boundaries during each wakeup of $s$
\State Report the result to $p$ when skill gates and \texttt{done\_when} are satisfied
\end{algorithmic}
\end{algorithm}

FairyClaw's single-step inference loop supports this implementation, but it is not the main abstraction. Its role is to provide clean intervention boundaries: after one LLM decision and its tool effects, the runtime persists the turn, updates skill state, and schedules a follow-up event if a gate remains open. This lets Formal Skill hooks observe every step without being hidden inside an opaque long-running planner loop. Dynamic routing, single-step wakeups, and parent-child collaboration are therefore runtime services for implementing Formal Skill.

\subsection{Case Study: \textsc{CodeRepairOps} as a Formal Skill}

\textsc{CodeRepairOps} instantiates Formal Skill for code repair and test-driven debugging. It is intentionally not a monolithic ``repair'' tool. The action plane exposes four structured tools for evidence collection, unified patching, verification, and artifact writing; the control plane uses hooks to filter those tools, inject phase-local guidance, validate calls, and keep the sub-session alive until gates pass. The implementation recognizes five phase labels: \texttt{reproduce}, \texttt{diagnose}, \texttt{patch}, \texttt{verify}, and \texttt{report}. Its injected default workflow is shorter, \texttt{reproduce -> patch -> verify -> report}; \texttt{diagnose} is an optional compatibility phase for additional evidence or contextual patching. The session state remains small: phase, verification status, failure signature, produced and required artifacts, gate-failure reasons, and last-tool metadata.

\begin{figure}[htbp]
\centering
\includegraphics[width=\linewidth]{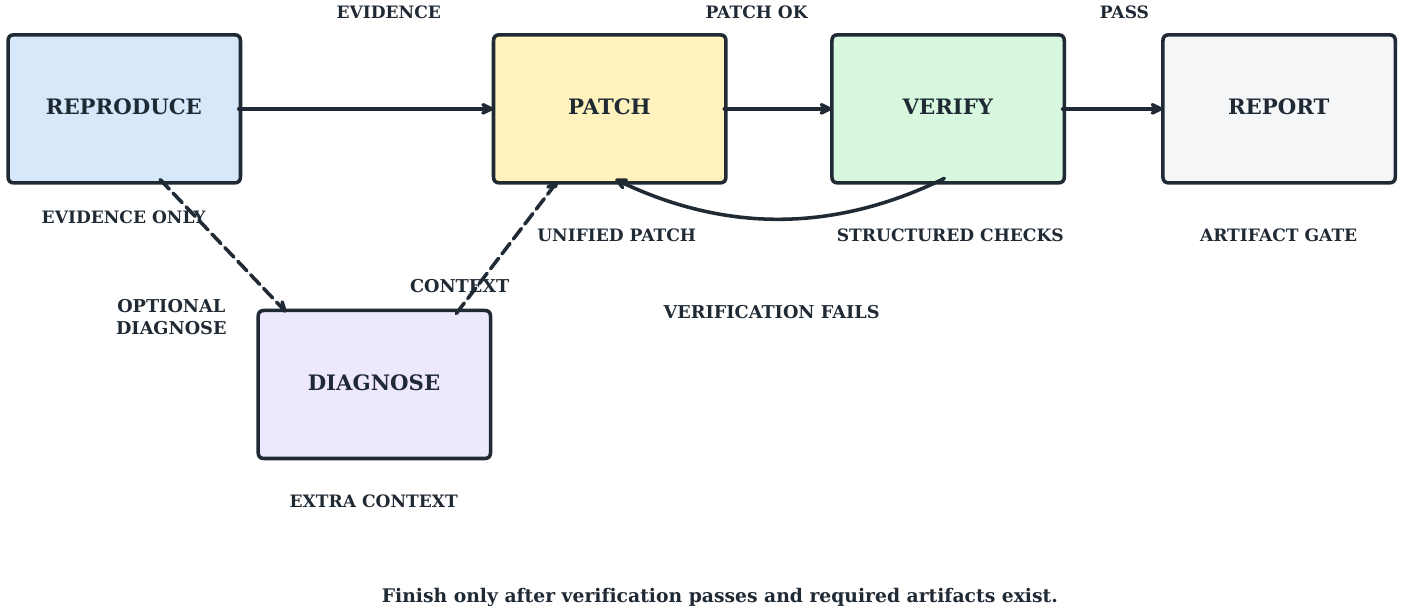}
\caption{\textsc{CodeRepairOps} state vocabulary and main transitions. In the current executor, evidence collection from \texttt{reproduce} may advance directly to \texttt{patch}; \texttt{diagnose} remains a routed phase for extra evidence or contextual patching.}
\label{fig:coderepair-state}
\end{figure}

Table~\ref{tab:coderepair-state} shows how the phase field is consumed before each model call. This is the main token-efficiency mechanism: the model receives only the tools and constraints relevant to the next decision, while the complete repair policy remains in executable code.

\begin{center}
\captionsetup{type=table}
\caption{\textsc{CodeRepairOps} phase policy. Tool names are abbreviated for layout; each maps to the corresponding \texttt{repair\_*} tool.}
\label{tab:coderepair-state}
\small
\setlength{\tabcolsep}{3pt}
\begin{tabular}{@{}p{0.14\linewidth}p{0.21\linewidth}p{0.58\linewidth}@{}}
\toprule
Phase & Visible actions & Hook policy and transition \\
\midrule
\texttt{reproduce} & collect evidence & Only evidence collection is visible. Command or log evidence updates the failure signature and advances to \texttt{patch}. \\
\texttt{diagnose} & collect, patch & Routed optional phase. Evidence collection and unified patching are visible; generic mutation tools remain hidden. \\
\texttt{patch} & collect, patch & Patching must use the unified patch tool; guards reject protected paths, missing hunks, append-only edits on existing files, and oversized changes. Success advances to \texttt{verify}. \\
\texttt{verify} & verify & Verification accepts structured checks. Passing checks set \texttt{verification\_passed} and advance to \texttt{report}; failures return to \texttt{patch}. \\
\texttt{report} & write artifacts, done & Artifact writing and completion are visible. Completion is blocked until verification has passed and required artifacts inferred from \texttt{done\_when} exist. \\
\bottomrule
\end{tabular}
\end{center}

\begin{algorithm}[H]
\caption{\textsc{CodeRepairOps} execution}
\label{alg:coderepair-execution}
\footnotesize
\begin{algorithmic}[1]
\Require Sub-session $s$ and repair state $r$
\State \texttt{before\_llm\_call} filters visible tools by $r.\texttt{phase}$ and injects phase-local workflow guidance
\State The model chooses one visible repair action
\State \texttt{before\_tool\_call} checks paths, patch shape, command policy, and verification checks
\State Execute the selected repair action with its Python executor
\State Update phase, verification status, produced artifacts, and gate reasons
\If{verification or artifact gates remain open}
  \State Schedule a follow-up wakeup for $s$
\Else
  \State Allow completion and return the repair report
\EndIf
\end{algorithmic}
\end{algorithm}

The injected workflow message is intentionally compact: current phase, visible tools, fixed workflow order, no-completion-without-verification rule, required-artifact reminders inferred from \texttt{done\_when}, patch-only-through-\texttt{repair\_apply\_unified\_patch}, and verification checks shaped as \texttt{\{name,type,args\}}. This keeps procedural knowledge in executable state and hooks rather than in a long natural-language repair manual.

\begin{center}
\captionsetup{type=table}
\caption{\textsc{CodeRepairOps} hook program. Each hook is a small deterministic boundary program around a single model or tool step.}
\label{tab:coderepair-hooks}
\footnotesize
\setlength{\tabcolsep}{3pt}
\begin{tabular}{@{}p{0.24\linewidth}p{0.32\linewidth}p{0.36\linewidth}@{}}
\toprule
Hook stage & Checks or payload rewrite & State-machine effect \\
\midrule
\texttt{before\_llm\_call} & Loads repair state, infers required artifacts from \texttt{done\_when}, filters visible tools by phase, and appends a user-role workflow message. & Does not advance the phase. It narrows the model's next action space so invalid actions are usually unreachable. \\
\texttt{after\_llm\_response} & If the model emits no tool call, forces evidence collection in \texttt{reproduce}, forces verification in \texttt{verify}, or schedules a report-stage follow-up when gates remain unresolved. & Prevents silent completion and re-enters the loop at the same phase with an explicit required action. \\
\texttt{before\_tool\_call} & Parses tool arguments. For patch calls, checks target existence, protected path patterns, and required hunk markers. For verification, requires a non-empty check list. & Blocks malformed calls before side effects. Protected edits are redirected to an artifact message; invalid patch or verification calls force another turn. \\
\texttt{after\_tool\_call} & Records last tool metadata, applies phase updates after patch, verification, or artifact writing, evaluates verification and artifact gates, and publishes follow-up events when needed. & Advances \texttt{patch -> verify}, \texttt{verify -> report} on pass, \texttt{verify -> patch} on fail, and keeps \texttt{report} open until completion gates pass. \\
\bottomrule
\end{tabular}
\end{center}

The same three limitations of prompt skills appear concretely here. Instead of asking the model to follow a repair order, \textsc{CodeRepairOps} changes tool visibility by phase. Instead of asking the model not to edit protected files, the patch tool and pre-tool hook reject unsafe edits. Instead of asking for verification before completion, after-response and after-tool hooks keep the session alive until verification and artifact gates pass. The result is not a black-box repair command but a compact executable protocol distributed across JSON schemas, Python executors, hooks, and state.

\section{Experiments and Analysis}

\subsection{Evaluation protocol}

We evaluate the Formal Skill approach as implemented in FairyClaw on the independent Harness-Bench suite \cite{harnessbench}. Harness-Bench evaluates agents in real workspaces with task fixtures, executable oracles, and usage accounting. From the benchmark repository, we use 75 tasks for this study. The evaluation compares FairyClaw, Moltis \cite{moltis}, NullClaw \cite{nullclaw}, ZeroClaw \cite{zeroclaw}, Hermes \cite{hermes_docs,hermes_architecture}, and OpenClaw \cite{openclaw_runtime,openclaw_tools} with two model families, \texttt{gpt-5.4} and \texttt{gemini-3.1-pro}.

For each task, the harness creates a fresh sandbox workspace, copies task fixtures into \texttt{workspace/in}, renders the task prompt with workspace-local paths, and launches the target agent through a harness adapter. During execution, a local usage proxy is injected through environment variables and placed between the agent runtime and its upstream model endpoint. The proxy records every model request and response into \texttt{requests.jsonl} and \texttt{responses/*.json}, including provider, response model, normalized input/output/cache/total token fields, assistant messages, and tool calls. For process grading, Harness-Bench reconstructs a compact call chain from these proxy records: it removes system messages, keeps user/assistant/tool-call content, tracks main-agent and sub-agent histories separately, skips context-compression calls, and emits only incremental conversation deltas plus the corresponding assistant response. Token totals are aggregated from the complete proxy log rather than from the shortened trace.

Scoring then combines executable task outcomes with process evidence. Each task provides an oracle module that inspects the final workspace and returns an \texttt{outcome\_score}; tasks that require semantic judgment may additionally attach an LLM-based quality score. The reconstructed call chain is passed to a rubric grader that scores tool-use appropriateness, consistency or flow coherence, robustness or error handling, and a binary security gate. The final benchmark score is computed as \texttt{combined = outcome\_effective × process\_effective × security}. The reported metrics are this combined score, total token usage, and average tokens per completed task. Our analysis asks three questions aligned with the paper's mechanism: whether FairyClaw is competitive on overall score, whether it improves token efficiency, and whether routed Formal Skills help on tasks that require controlled evidence, action, verification, and recovery.

\subsection{Overall score and token efficiency}

Table~\ref{tab:overall-results} summarizes the evaluated runs. On \texttt{gpt-5.4}, FairyClaw achieves the strongest combined score among the compared agents, 0.746 over 75 tasks, while also using the fewest total tokens: 3.51M total and 46.8K per task. This is a favorable score--cost point: Moltis and Hermes are close in score, but use 6.29M and 7.28M tokens respectively; OpenClaw, NullClaw, and ZeroClaw score lower in this snapshot.

\begin{table}[htbp]
\centering
\caption{Harness-Bench comparison across evaluated agent harnesses.}
\label{tab:overall-results}
\footnotesize
\begin{tabular}{llrrrr}
\toprule
Agent & Model & Tasks & Combined $\uparrow$ & Total tok. $\downarrow$ & Avg. tok. $\downarrow$ \\
\midrule
FairyClaw & \texttt{gpt-5.4} & 75 & 0.746 & \textbf{3.51M} & 46.8K \\
Moltis & \texttt{gpt-5.4} & 75 & 0.744 & \textbf{6.29M} & 83.8K \\
Hermes & \texttt{gpt-5.4} & 75 & 0.740 & \textbf{7.28M} & 97.0K \\
OpenClaw & \texttt{gpt-5.4} & 75 & 0.693 & \textbf{6.03M} & 80.4K \\
NullClaw & \texttt{gpt-5.4} & 75 & 0.685 & \textbf{8.16M} & 108.9K \\
ZeroClaw & \texttt{gpt-5.4} & 75 & 0.587 & \textbf{6.09M} & 81.2K \\
\midrule
OpenClaw & \texttt{gemini-3.1-pro} & 75 & 0.689 & \textbf{6.58M} & 87.7K \\
NullClaw & \texttt{gemini-3.1-pro} & 75 & 0.645 & \textbf{12.70M} & 169.3K \\
Moltis & \texttt{gemini-3.1-pro} & 75 & 0.643 & \textbf{6.27M} & 83.6K \\
FairyClaw & \texttt{gemini-3.1-pro} & 75 & 0.635 & \textbf{3.84M} & 51.2K \\
Hermes & \texttt{gemini-3.1-pro} & 75 & 0.622 & \textbf{6.21M} & 82.8K \\
ZeroClaw & \texttt{gemini-3.1-pro} & 75 & 0.506 & \textbf{4.86M} & 64.8K \\
\bottomrule
\end{tabular}
\end{table}

Across both model families, FairyClaw's automatic context-compression mechanism before reaching the maximum context window was disabled. The observed token reduction therefore comes from routed Formal Skills, compact phase guidance, and narrower tool exposure rather than from aggressive transcript compression. On \texttt{gemini-3.1-pro}, FairyClaw is not the highest aggregate-scoring agent, but it remains competitive while preserving the strongest token-efficiency profile. It uses 3.84M total tokens, compared with 4.86M for ZeroClaw, 6.21M for Hermes, 6.27M for Moltis, 6.58M for OpenClaw, and 12.70M for NullClaw.

\subsection{Skill-relevant task behavior}

The aggregate comparison is only one view of Formal Skill. Its mechanism should matter most on procedural tasks where the agent must avoid unsafe shortcuts and premature completion. Table~\ref{tab:skill-task} reports one such code-debugging task from the benchmark: the workspace contains \texttt{buggy\_code.py} with five layered bugs, only the current layer is exposed in each round, and the agent must edit the file, validate it, and stop after the current layer passes. FairyClaw obtains the strongest combined score in both model families: 0.865 with \texttt{gpt-5.4} and 0.843 with \texttt{gemini-3.1-pro}. This is consistent with \textsc{CodeRepairOps}' purpose: make the repair loop executable through phase-specific tool visibility, patch guards, verification checks, and completion gates.

\begin{table}[htbp]
\centering
\caption{Representative code-debugging task results. This task stresses evidence collection, workspace modification, verification, and recovery.}
\label{tab:skill-task}
\footnotesize
\begin{tabular}{llrrr}
\toprule
Agent & Model & Combined $\uparrow$ & Total tok. $\downarrow$ & Rounds $\downarrow$ \\
\midrule
FairyClaw & \texttt{gpt-5.4} & 0.865 & 193.0K & 48 \\
NullClaw & \texttt{gpt-5.4} & 0.858 & 193.0K & 21 \\
Hermes & \texttt{gpt-5.4} & 0.807 & 750.5K & 99 \\
Moltis & \texttt{gpt-5.4} & 0.793 & 306.8K & 24 \\
OpenClaw & \texttt{gpt-5.4} & 0.730 & 349.1K & 24 \\
ZeroClaw & \texttt{gpt-5.4} & 0.700 & 207.8K & 24 \\
\midrule
FairyClaw & \texttt{gemini-3.1-pro} & 0.843 & 209.3K & 54 \\
Hermes & \texttt{gemini-3.1-pro} & 0.836 & 330.2K & 89 \\
OpenClaw & \texttt{gemini-3.1-pro} & 0.706 & 290.5K & 23 \\
NullClaw & \texttt{gemini-3.1-pro} & 0.692 & 293.0K & 29 \\
Moltis & \texttt{gemini-3.1-pro} & 0.657 & 154.8K & 19 \\
ZeroClaw & \texttt{gemini-3.1-pro} & 0.537 & 138.4K & 21 \\
\bottomrule
\end{tabular}
\end{table}

These task-level results should not be interpreted as a claim that every task is solved by \textsc{CodeRepairOps}. Rather, they identify the regime in which Formal Skill is most visible: the task has a meaningful procedure and the procedure has unsafe shortcuts. FairyClaw may spend more rounds than some baselines, but those rounds are governed by Formal Skill routing and hook-controlled state transitions rather than by an unstructured transcript. Its lower token consumption is evidence that this longer trajectory is not driven by verbose prompt accumulation, but by compact state-conditioned control.

\subsection{Analysis}

The experiments show three main results. First, FairyClaw is competitive on overall Harness-Bench score: across the six evaluated harnesses, it ranks third by mean score over the two model families and is the top-scoring harness in the \texttt{gpt-5.4} group. Second, FairyClaw uses substantially fewer tokens than the compared systems. Because context compression was disabled during these runs, the token reduction is attributable to Formal Skill routing, compact phase guidance, and narrower tool exposure rather than transcript summarization. Third, FairyClaw performs especially well on the representative code-debugging task, where \textsc{CodeRepairOps} makes evidence collection, patching, verification, and completion gates explicit. Together, these results suggest that Formal Skill improves agent efficiency while preserving, and in some procedural tasks improving, task performance.

\section{Conclusion}

We presented \textbf{Formal Skill}, a runtime-native way to turn reusable agent skills from long prompt instructions into programmable protocols with JSON schemas, reliable executors, hooks, routing metadata, and skill-local state. FairyClaw implements this concept as an event-driven agent runtime, and \textsc{CodeRepairOps} shows how it becomes a concrete workflow for code repair. On Harness-Bench, FairyClaw achieves competitive overall scores, ranks first in the \texttt{gpt-5.4} group and third by mean score across the six evaluated harnesses, and uses substantially fewer tokens. These results support the view that agent skills should be token-efficient, runtime-governed protocols rather than natural-language documents alone.

\end{document}